\documentclass[a4paper,12pt]{article}
\usepackage[utf8]{inputenc}
\usepackage[english]{babel}
\usepackage{authblk}
\usepackage{adjustbox}
\usepackage{graphicx}
\usepackage{mathptmx}
\usepackage{tabularx}
\usepackage{enumitem}
\usepackage[title]{appendix}
\usepackage[singlespacing]{setspace}
\usepackage[headheight=1in,margin=1in]{geometry}
\usepackage{fancyhdr}
\usepackage{lipsum}
\usepackage{caption}
\usepackage{hyperref}
\makeatletter
\def\@maketitle{%
  \newpage
  \begin{center}%
  \let \footnote \thanks
    {\LARGE \@title \par}%
    \vskip 1.5em%
    {\large
     \lineskip .5em%
      \begin{tabular}[t]{c}%
        \@author
      \end{tabular}\par}%
    \vskip 1em%
    {\large \@date}%
  \end{center}%
  \par
  \vskip 0.1em}
\makeatother

\graphicspath{{images/}}

\title{Multiclass Classification of Policy Documents with Large Language Models}
\author{Erkan Gunes, Christoffer Florczak}
\affil{Aalborg University}
\date{October 12, 2023}

\begin{document}

\doublespacing
\maketitle
\thispagestyle{fancy}

\begin{abstract}
Classifying policy documents into policy issue topics has been a long-time effort in political
science and communication disciplines. Efforts to automate text classification processes for social science research purposes have so far achieved remarkable results, but there is still a large room for progress. In this work, we test the prediction performance of an alternative strategy, which requires human involvement much less than full manual coding. We use the GPT 3.5 and GPT 4 models of the OpenAI, which are pre-trained instruction-tuned Large Language Models (LLM), to classify congressional bills and congressional hearings into Comparative Agendas Project’s 21 major policy issue topics. We propose three use-case scenarios
and estimate overall accuracies ranging from \%58-83 depending on scenario and GPT model
employed. The three scenarios aims at minimal, moderate, and major human interference,
respectively. Overall, our results point towards the insufficiency of complete reliance on GPT
with minimal human intervention, an increasing accuracy along with the human effort exerted,
and a surprisingly high accuracy achieved in the most humanly demanding use-case. However, the superior use-case achieved the \%83 accuracy on the \%65 of the data in which the two
models agreed, suggesting that a similar approach to ours can be relatively easily implemented
and allow for mostly automated coding of a majority of a given dataset. This could free up resources allowing manual human coding of the remaining \%35 of the data to achieve an overall
higher level of accuracy while reducing costs significantly.

\end{abstract}

\section*{Introduction}

Text is a highly valuable source of information for political scientists as political actions, debates and outcomes are often the main subject in a wide array of different documents for instance newspaper articles, parliamentary and bureaucratic records, social media posts, press releases, court documents and political manifestos (Grimmer \& Stewart, 2013; Wilkerson \& Casas, 2017). Due to the inherently unstructured nature of textual data it has traditionally been analyzed with primarily qualitative methods, which has limited the amount of data that could be processed. The rise of computational social science has enabled researchers to train algorithms for their particular use cases with off-the-shelf methods of machine learning to encompass a very high volume of textual data, thus allowing a seemingly revolutionary efficiency increase and expanding the possibilities of testing observable implications of theories in many branches of political science. (Lazer et al., 2009; Gentzkow, Kelly \& Teddy, 2019)  

The increase in valuable computational tools does not come without costs (Denny and Spirling, 2018). One of the most discussed issues is the error rate of computational language models which has, perhaps due to this focus, been drastically improved upon through technological advances such as neural networks, transformers, and Large Language Models (LLM) (Bosley et. al., 2023). In this paper, we therefore approach the literature from the vantage point of other less debated issues that are hindering the wide adoption of computational text techniques and test if the recent advance of instruction-tuned LLMs, and particularly OpenAI’s GPT (OpenAI, 2023) models, provide sufficient answers to these challenges. The first challenge is that training one’s own algorithm requires both theoretical and practical expertise which puts the entry barrier to adoption relatively high. The costs of adopting the technical aspects of computational text analysis approaches are unlikely to be evenly distributed as has historically been the case during other major methodological transitions in social sciences. In an academic reality with increasing demands and time constraints this could in turn prevent some parts of the scholarly community from adopting these newer tools. The second challenge is that training context specific algorithms require substantial text corpora, and in some cases also require manual hand coding of a large amount of text. This can make the use of computational methods unfeasible or impractical in cases where the text amounts needed for a given analysis are more than one could reasonably handle by hand-coding but at the same time insufficient for training an algorithm from scratch. The third challenge is that many conventional computational methods for text classification require multiple steps of text pre-processing, which requires extensive documentation to ensure the reproducibility of the classification output. Additionally, many computational approaches are highly sensitive to text pre-processing choices, and limiting the amount of possible variability in those choices could improve consistency across research projects.   

At face value, the recent development of instruction-tuned LLMs, such as OpenAI’s GPT models (Radford et. al., 2019), might offer an answer to those challenges. First, OpenAI provides a simple interface to interact with their models, thereby substantially decreasing the amount of technical expertise needed to use the algorithm. In practice, it reduces technical requirements by moderately shifting the focus from programming to prompt-engineering. Furthermore, access to OpenAI’s API makes projects highly scalable and might allow the intuitive understanding to be developed using the interface and then translated to the programming environment. Secondly, the pre-trained nature of the GPT models means that use by the individual researcher would be akin to zero-shot learning (Radford et. al., 2019), as the model requires no training to be used. Consequently, the quality of the results derived from the model is therefore much less dependent on the amount of data the individual researcher puts into it. However, implementing any language model in an academic context on face value without tests of its performance is ill-advised at best. The aim of this paper is therefore to provide initial indications of how some of the most widely used instruction-tuned LLMs might perform on a concrete multi-class classification task with relevance for political science.

To provide such a test we use data on congressional bills and congressional hearings and instruct the GPT models to classify the bills according to the coding scheme from the Comparative Agendas Project. This provides us with the opportunity to evaluate the performance of recent GPT models to the ‘gold standard’ of human coding. To get a sense of how the improvements in the algorithm between versions affect the model’s ability to perform classification tasks we carry out tests with both GPT 3.5 and GPT 4. Furthermore, this allows us to compare agreement rates between the two models. To situate and illustrate researchers’ use of GPT, we develop three scenarios or use-cases and report results for each scenario. In the first use-case, researchers rely only on GPT for classification with no human intervention. In the second use-case, humans assist with cleaning and structuring. In the third case, GPT models are combined to achieve a high accuracy score for a subset of the data, and humans assist with cleaning, structuring and manual coding of remaining data. This allows us to evaluate how the algorithm performs along varying degrees of human effort exerted.

Our results suggest that while the GPT models perform surprisingly well on multi-class classification tasks, the accuracy is still some way from rivaling the performance seen in similar classification tasks from the state of the art custom-tailored approaches. Our comparison between the GPT 3.5 and GPT 4 models suggests a significant increase in accuracy between the two versions of the algorithm, signifying that LLM’s such as GPT may reach efficiency levels appropriate for use in the social sciences in the future. Comparing the two models, we see a roughly \%65 agreement between the classifications. If we isolate to only the cases in which the two models agree, the overall accuracy score is above \%80, thereby rivaling many state-of-the-art approaches to classification in political science. Evaluating our use-cases, we conclude that complete reliance on GPT is suboptimal and should be avoided, while significant cost reduction and highly promising accuracy results can be achieved when combining machine and human effort to the highest degree. We conclude the paper by discussing prompt engineering, field specificity, combining models, and the possibility of fine-tuning the GPT models in the future for use in political science. 

\section*{Large Language Models and Social Science Research}

Large Language Models (LLMs) are distinguished by the amount of data they are trained on, the scale of the neural network underlying them, and the scope of tasks they are aimed to achieve. Computational social science research on text classification so far has mainly progressed with the use and development of custom natural language processing models specializing on specific tasks and trained on data to optimize the performance on the specific task. LLMs are foundation models which aim to imitate natural language generation by humans; therefore they have the potential to perform all kinds of specific text analysis and generation tasks specialized models are trying to implement, though possibly with varying levels of performance across different tasks (Kocon et al., 2023). This flexibility in LLMs use cases in the context of text annotation is one of the main advantages of LLMs compared to more conventional computational text annotation methods.  

LLMs could be divided into two types: base LLMs and instruction-tuned LLMs. The former one is basically a model that predicts the next token given the provided text. Instruction-tuned LLMs use next token prediction mechanism to generate responses to user prompts. They are developed using examples of prompt and response pairs as training data, and could be further refined using reinforcement learning from human feedback. The use of LLMs for various NLP tasks have been surging in the last few years thanks to the introduction of powerful models such as BERT and Generative Pre-Trained Transformer (GPT), and it has started gaining more momentum with the introduction of instruction-tuned LLMs and user interfaces that enable conversational interaction with instruction-tuned LLMs, such as ChatGPT. 

Text classification using instruction-tuned LLMs differs from conventional unsupervised approaches like topic models and custom supervised models. While topic models are traditionally fully unsupervised \footnote{For recent developments in semi-supervised topic modelling in political science, see: 
  Eshima, Imai and Sasaki, 2023. 
} and identify latent topics rather than known classes, instruction-tuned LLMs can be guided towards known target classes through instructions passed as a prompt to the model. Compared to custom supervised models, LLMs require no labeled training data, instead relying on pre-trained knowledge to infer the label of a given text based on the context and instructions provided. This reduces the need for expensive and time consuming data labeling processes, which are often necessary for traditional supervised learning. Furthermore, instruction-tuned LLMs can quickly adapt to new classification tasks without the need for retraining from scratch. This makes them highly flexible and versatile across a wide range of applications. However, the efficacy of instruction-tuned LLMs can be contingent upon the clarity and specificity of the instruction prompts. In scenarios where the distinction between classes is subtle, a well-crafted prompt becomes crucial. Nonetheless, this approach shows the power of leveraging vast amounts of pre-existing knowledge to perform text classification without extensive fine-tuning or labeled datasets.

Research on the efficacy of instruction-tuned LLMs for various social science research tasks have been growing rapidly, especially since the introduction of ChatGPT in late 2022. Some remarkable examples include the use of OpenAI's GPT models for simulating survey responses using sociodemographic information about potential survey respondents (Argyle et al, 2023), measuring latent ideology of politicians (Wu et al., 2023) and detecting news pieces with misinformation or disinformation  (Caramancion, 2023). Recently, we have seen an influx of studies evaluating the capabilities of those models in performing diverse text annotation tasks such as topic detection, stance detection, and frame detection in tweets (Gilardi, Alizadeh and Kubli; 2023), sentiment analysis in tweets (Zhu et al, 2023) and classification of Twitter accounts by political affiliation (Törnberg, 2023). A prevailing conclusion in those studies is that the 
more difficult the text annotation problem, e.g. larger number of classes, fuzzy semantic boundaries etc., the lower the performance of GPT models compared to the state of the art model(s) for the specific text annotation task (Kocon et al, 2023). However, in most scenarios, they were able to achieve sufficiently good performance with a zero shot learning strategy which requires much less overall effort than human labelling or computational labelling using more conventional methods. In this study, we test text classification capabilities of the two most advanced instruction-tuned LLMs released by the OpenAI, i.e. GPT 3.5 and GPT 4, on a relatively more complex classification task. We use those two models to classify policy documents into policy issue topic categories, which is a multiclass classification task with more than 20 distinct classes. In the next section, we elaborate on the data and the procedures we followed in our tests.\\

\section*{Data and Research Setup}

We use two large datasets from the Comparative Agendas Project's database which include information about the issue topic category of legislative policy documents and events. The first dataset contains policy bill titles from the US Congress and issue topic labels assigned by human research assistants. The second one contains congressional hearing descriptions and issue topic labels assigned by human research assistants. Both datasets use the CAP topic scheme, which contains 21 major topics such as macroeconomics, health, agriculture, etc., and numerous subtopics under each major topic. The congressional bills dataset has an additional category , i.e. private bills, which we named 'other' in our tests. The extensive use of these two datasets in previous computational social science research to examine various text classification approaches allows us to benchmark the performance of our method against other strategies. Summary information about those data could be seen in Table 1 . 

\begin{table}[h]
    \caption{Description of Data}
    \centering
    \small
    \begin{tabularx}{\textwidth}{|X|c|c|X|}
    \hline
    \textbf{Dataset} & \textbf{N} & \textbf{t} & \textbf{Example Text} \\
    \hline
    Congressional Bills & 468,438 & 1947-2016 & 
    \begin{itemize}[noitemsep,nosep,left=0pt]
    \item A bill to require the Secretary of Homeland Security and the Secretary of State to accept passport cards at air ports of entry and for other purposes.
    \item To amend the Internal Revenue Code of 1986 to provide a tax credit for the costs of college textbooks.
    \item A bill to amend the Small Business Act to direct the task force of the Office of Veterans Business Development to provide access to and manage the distribution of overseas excess or surplus property to veteran-owned small businesses.
    \end{itemize}  \\
    \hline
    Congressional Hearings & 102,151 & 1946-2020 & 
    \begin{itemize}[noitemsep,nosep,left=0pt]
    \item Scope of Soviet activity in the US.
    \item Tax treatment of recycling of solid waste.
    \item World petroleum outlook for 1982.
    \end{itemize}  \\
    \hline
    \end{tabularx} 
\end{table}

We use two slightly different learning strategies to test the capabilities of two pretrained LLMs from OpenAI, the GPT3.5 and the GPT4 algorithms respectively. In our tests with the GPT 3.5 we use a fully zero-shot learning strategy, while we use a mostly zero shot learning strategy in our experiments with the GPT 4 algorithm. In the fully zero-shot learning situation, we do not show any examples of text and label pairs from the dataset for any of the classes. In the mostly zero-shot situation, we show a few text and label pairs for two of the classes. An alternative strategy, which we do not use in this work, would be few shot learning in which we would show a few examples for each candidate labels in our prompts. Studies indicate few shot learning strategy could  significantly improve LLM performance (Brown et al., 2020), however,  we limit the use of that strategy to classes on which the GPT 4 algorithm was struggling to get right in our small non-systematic tests which we did before running the large scale tests whose results we present in the next section. We couldn't use that strategy for all classes, because providing even a couple labeled examples for each of the 21 classes would exceed the context length limits, especially with the congressional bill titles which on average have longer token length than hearings descriptions. However, with token limits for LLMs having increased since the time of our tests, few-shot learning strategies with LLMs have become more feasible. Therefore, zero-shot learning provides a more convenient and economical approach for classifying documents into the CAP major topics, as no hand-labeled examples are required. 

\begin{figure}[htp]
\centering
\includegraphics[width=15cm]{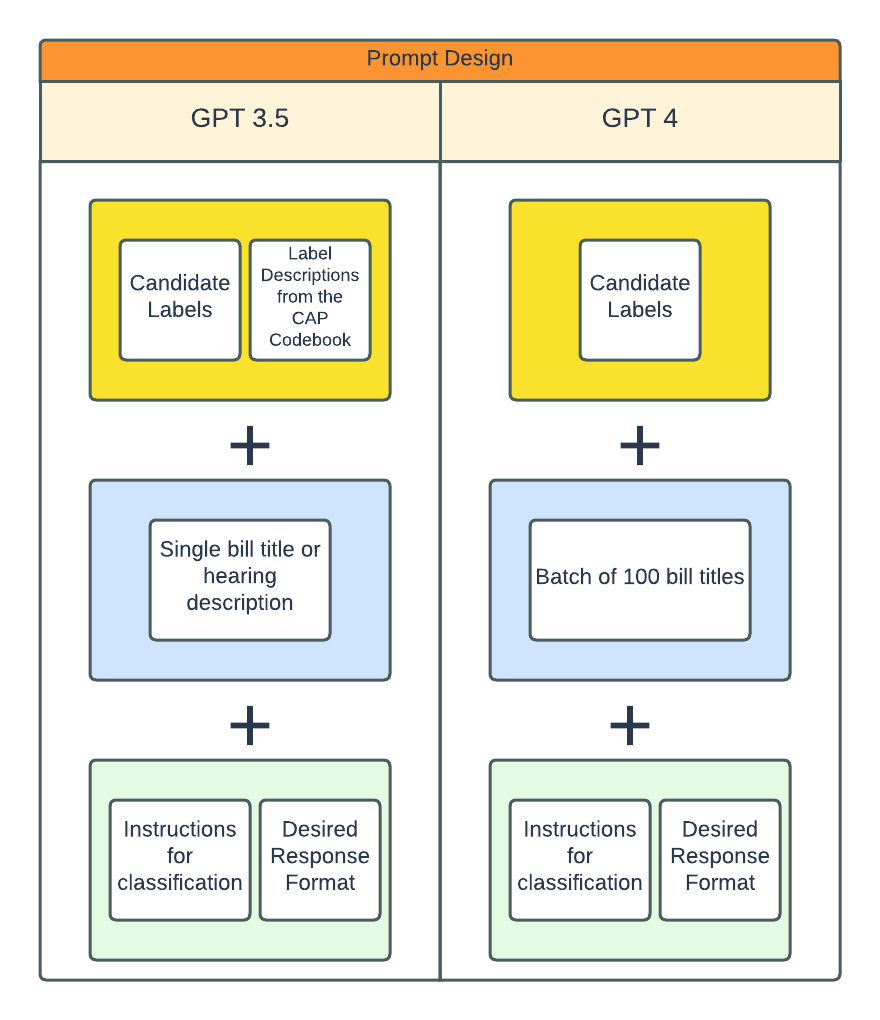}
\caption{Prompt Design}
\label{fig:image}
\end{figure}

We use the prompt structures depicted in Figure 1 in our text classification tests with the OpenAI's LLMs. The cost per token and context length differences between the two models were the main reasons why we used two different prompt structures. Since our main focus in this work is not comparing the two models, but to offer an alternative approach to existing text classification approaches in computational social science, we do not pay special attention to the alignment between prompts we use in our tests with these two different models. However, in spite of the limited context we were able to provide in the GPT 4 prompts and asking the algorithm to classify 100 titles in a single prompt, we observed a considerable improvement when we used the GPT4 algorithm.

GPT 3.5 model is much cheaper than the GPT 4 model. At the time we conducted these tests, the former models' cost per 1K token was \$0.002 and \$0.03 for the latter model. In our experiments with GPT 3.5, we included a single bill title or hearing description in each prompt, along with the class labels, their full descriptions from the CAP master codebook and instructions about the classification task. For GPT 4, with higher token limits and higher cost per token, we include a batch of multiple titles or hearings in each prompt, along with just the candidate class labels and some instructions about the nuances of some issue topics. In GPT 4 prompts, we also included a few example titles from the private bills and government operations category. Private bills have a very standard structure, where each title begins with the phrase "for the relief of". A researcher working with the congressional bills data will quickly realize that structure. We showed two examples from that category in our prompts. In our small scale experiments, we also realized the GPT 4 algorithm was struggling with the government operations class and decided show an example of that class, too. We also put some information to help the GPT 4 algorithm on issues that could be associated with multiple classes such as abortion, veterans affairs. In the CAP scheme, abortion related issues are classified under the civil rights topic, but in our small scale experiments, we realized the GPT 4 algorithm sometimes associated that issue with the health category. The same problem was observed for the veterans affairs issue, where the GPT 4 algorithm sometimes put veterans affairs related issues under the social welfare topic, while the CAP scheme puts it under the defense topic. Examples of full prompts we used with the GPT 3.5 and GPT 4 algorithms could be found in the Appendix A.  

To query the LLMs, we use OpenAI's API which provides convenient access to models like GPT 3.5 and GPT 4 through API calls in Python. We specifically use the gpt-3.5-turbo-0301 and the gpt-4-0314 models. The last four digits in model names refer to the dates the models were released. OpenAI later released updated versions of those models, which had not been released by the time we did our tests. OpenAI's API allows controlling some model parameters such as the temperature parameter which control the randomness in model output. That parameter ranges from 0 to 1, with lower values resulting in more conservative, deterministic responses. We set the temperature parameter to 0 for both models to obtain almost deterministic and reproducible results. Although those models are not perfectly deterministic at temperature 0, they produce near-identical results across repeated runs. We generate responses from the models in a pre-defined format, asking them to output just the predicted class label for each input observation. However, GPT 3.5 was not always successful in terms of generating the response in the desired format, while GPT 4 always gave the response in the desired format. \\

\section*{Results}

We present the results using three different scenarios, which correspond to three different ways researchers may use the LLM generated classification output. In each of those scenarios, we have increasing levels of human involvement in the text classification process. Human-computer collaboration has been suggested as a strategy to improve text classification productivity and quality (Loftis and Mortensen, 2020), and here we use these scenarios to show how human-computer collaboration can also improve the performance of text classification with LLMs. In the first scenario, we report classification performance metrics where the researchers will use the full LLM output without touching the predictions regardless of whether they belong to the set of candidate labels. In the second scenario, researchers will exclude the text with predicted labels which do not exist within the candidate labels. In the third scenario, researchers will use a label prediction only when the two algorithms agree.

\subsection*{Scenario 1: Using Untouched LLM Predictions}

In the first scenario, we present results based on the assumption that researchers will not touch the output from the LLM models. They will utilize the predicted labels as they are, even if they do not exist in the candidate labels set. Our aim in presenting that scenario is to demonstrate the performance achievable with minimal human intervention in the classification process. As such, any classifications with labels that do not appear in the candidate labels set are treated as incorrectly coded in that scenario.\\


\begin{table}[h!]
\centering
\caption{Overall accuracy performance of GPT models when all label predictions are included in the evaluation}
\begin{tabular}{|l|l|p{2cm}|l|l|p{2cm}|}
\hline
\textbf{Model} & \textbf{Dataset} & \textbf{Sample} \newline \textbf{Size} & \textbf{Accuracy} & \textbf{F1 Score} & \textbf{Weighted F1 Score} \\
\hline
GPT 3.5 Turbo  & Bills            & 11,300& 0.63& 0.55& 0.63\\
\hline
GPT 3.5 Turbo  & Hearings         & 11,011      & 0.59& 0.56& 0.61\\
\hline
GPT 4          & Bills            & 11,300& 0.69              & 0.60& 0.70 \\
\hline
\end{tabular}
\end{table} 

Table 2 shows the performance metrics of different GPT models under that scenario. It specifically highlights three  performance indicators for models: accuracy, F1 Score, and weighted F1 Score. These metrics are essential as they provide an overall view of the model's ability to correctly predict and balance precision and recall, which respectively refer to how often the model is correct when it predicts a certain class and how well the model identifies all instances of each class.

We tested the GPT 3.5 Turbo model on the Congressional Bills dataset  and the Congressional Hearings dataset, while the GPT 4 model was tested only on the Congressional Bills dataset. The sample sizes for these datasets vary, with the Congressional Bills dataset having slightly larger sample sizes in both model instances. Starting with GPT 3.5 Turbo, on the congressional bills dataset, the model achieved an accuracy of 0.63, an F1 Score of 0.55, and a weighted F1 score of 0.63. When the same model was tested on a random sample from the congressional hearings dataset there was a moderate dip in accuracy to 0.59, but maintained an F1 Score of 0.56 and a slightly decreased weighted F1 score of 0.61. Conversely, when we evaluated the GPT 4 model using the Congressional Bills dataset, there was a marked improvement in performance. The model reached an accuracy of 0.69, outperforming the GPT 3.5 Turbo. Additionally, the F1 Score and weighted F1 score for GPT 4 on this dataset were 0.60 and 0.70 respectively. This suggests that while both models show competency in classifying data from the two datasets, the GPT 4 model exhibits superior performance on the Congressional Bills dataset when compared to its predecessor.

In Appendix B, Table 5 and Table 6 present the class-specific performance metrics for GPT 4 and GPT 3.5 with the congressional bills dataset. Overall, GPT 4 demonstrates comparable or superior performance in precision, recall, and F1 Score across a majority of the classes when juxtaposed against GPT 3.5. Particularly on the health, agriculture and private bills (Other) topics, both models exhibit high accuracy performance, with GPT 3.5 Turbo marginally leading on the health topic and the private bills topics . The immigration topic reveals a notable divergence in performance, where GPT 3.5 significantly overperforms GPT 4, especially in terms of F1 score. Though GPT 4 manifests heightened recall on both the environment and technology topics, this leads to some compromise in precision within the environment class. On the topics of defense and international affairs, GPT 4 overperforms GPT 3.5 by a large margin.

In Table 7, we present GPT 3.5 model's class-specific performance on the congressional hearings dataset. We do not observe any class with F1 score above 80 percent, while the health, agriculture, and energy topics achieve F1 scores close to that level. Some classes, including labor, education, and transportation, hover around the mid-70s in terms of F1 score, suggesting a relatively consistent performance across various topics in the dataset.

\subsection*{Scenario 2: Omitting Non-Matching Predicted Labels}

In the second scenario, we assume the researchers will filter out the predicted labels which do not match with any candidate labels from the topic scheme and responses which do not match the desired response format. For example, in our GPT 3.5 experiment with the congressional bills data, the predicted labels set has some labels such as 'veterans affairs' and 'tax policy', which are not among the candidate labels we provided in our prompts. In our experiments with the congressional hearings dataset, the GPT 3.5 model sometimes gave responses with explanations of the prediction or responses that have additional text to the label prediction. In that second scenario, we assume the researchers will put those kinds of predictions aside and label those texts manually. That will require a bit more human involvement in the process compared to the first scenario, but this will result in a considerable classification quality gain with minimal additional human effort in the process. We report the classification performance results for the machine labelled portion of the data in the below table. \\


\begin{table}[h!]
\centering
\caption{Overall accuracy performance of GPT models when the predicted labels that do not exist in the coding scheme provided in the prompt are excluded}
\begin{tabular}{|l|l|p{2cm}|l|l|p{2cm}|}
\hline
\textbf{Model} & \textbf{Dataset} & \textbf{Sample} \newline \textbf{Size} & \textbf{Accuracy} & \textbf{F1 Score} & \textbf{Weighted F1 Score} \\
\hline
GPT 3.5 Turbo  & Bills            & 10,662 & 0.67 & 0.59 & 0.66 \\
\hline
GPT 3.5 Turbo  & Hearings         & 10,059 & 0.64 & 0.61 & 0.65 \\
\hline
GPT 4          & Bills            & 11,300 & 0.69 & 0.60 & 0.70 \\
\hline
\end{tabular}
\end{table}


The GPT 3.5 model achieves an accuracy of 0.67, F1 score of 0.59, and a weighted F1 score of 0.66 when we test in with the congressional bill titles. The same model on the congressional hearings dataset achieves an accuracy of 0.64, an F1 score of 0.61 and a weighted F1 score of 0.65. The GPT 4 model's metrics for the Congressional Bills dataset remained same because it did not suffer from the same problem where GPT 3.5 made up labels that do not exist in the candidate labels set or did not generate the response in the desired format. 

\subsection*{Scenario 3: Trusting Only Mutual Predictions of GPT Models}

In the third scenario, we assume the researcher will only rely on the predictions where the two models agree. We have test results from both models for the congressional bills data, but not for the congressional hearings data. Thus, we only report results for the congressional bills data in that scenario. Table 4 shows the overall classification performance metrics for the machine labelled portion of the data. The two models agreed on approximately \%65 of the congressional bill titles. Within that subset of the random sample, the accuracy is significantly higher than the overall accuracy of both models in the other two scenarios presented earlier. While there is a much larger classification quality gain in that scenario, now the researcher will need to spend more human effort on the approximately \%35 of the data where the two models disagreed. \\


\begin{table}[h!]
\caption{Overall accuracy performance of GPT models when the titles on which the two models' predictions disagree are excluded}
\begin{adjustbox}{width=\columnwidth,center}
\begin{tabular}{l|l|l|l|l|l}
\textbf{Model} & \textbf{Dataset} & \textbf{Sample Size} & \textbf{Accuracy} & \textbf{F1 Score} & \textbf{Weighted F1 Score} \\
\hline
GPT 3.5 Turbo  & Bills            & 7,291& 0.83& 0.73& 0.83\\
\hline
GPT 4          & Bills            & 7,291& 0.83& 0.73& 0.83\end{tabular}
\end{adjustbox}
\end{table}


In that scenario, both GPT models achieve an accuracy of 0.83 for the congressional bills dataset. This is a significant improvement compared to the first and second scenarios where GPT 4, the better performing model, achieved an accuracy of 0.69. We observe similar improvement in F1 score and weighted F1 score metrics. GPT 4's F1 score in the third scenario went up to 0.73 from the 0.60 score in the firs and second scenarios, and the weighted F1 score went up to 0.83 from 0.70 in the previous two scenarios. This enhancement indicates that when both models agree on predictions, classification quality substantially increases. However, this approach also leaves \%35 of the titles on which the models diverged unaddressed.

Class-specific performance metrics also suggest a significant improvement over the other two scenarios. Now, there are ten topics with at least \%80 F1 score, and all topics except "international affairs" have at F1 scores above \%50. The 'private bills" category, which is the most common category in the random sample, as well as in the full dataset, has almost perfect accuracy.\\

\section*{Discussion}

In this paper we tested how the 3.5 and 4.0 versions of the GPT model performed several multiclass classification tests aiming at identifying a list of prespecified political topics commonly used in the CAP project in inherently political texts. We proposed three use-case scenarios and estimated overall accuracies ranging from \%58-83 depending on scenario and GPT model employed. The three scenarios aimed at minimal, moderate, and major human interference, respectively. Overall, our results point towards the insufficiency of complete reliance on GPT with minimal human intervention, an increasing accuracy along with the human effort exerted, and a surprisingly high accuracy achieved in the most humanly demanding use-case. However, the superior use-case achieved the \%83 accuracy on the \%65 of the data in which the two models agreed, suggesting that a similar approach to ours can be relatively easily implemented and allow for mostly automated coding of a majority of a given dataset. This could free up resources allowing manual human coding of the remaining \%35 of the data to achieve an overall higher level of accuracy while reducing costs significantly.

Our results are slightly less optimistic than other recent papers examining how GPT models might help political scientists. For instance, Wu and colleagues (2023) show that GPT 3.5 can be used to scale American politicians according to ideological orientation. Furthermore, Törnberg (2023) show that ChatGPT outperforms both crowd and Mturk coders on classifying whether tweets originate from democrats or republicans, suggesting that LLMs might sometimes be superior to frequently used human alternatives. Gilardi and colleagues (2023) similarly show that GPT outperforms coders at classifying tweets in six categories. Comparing our results with those of Törnberg and Gilardi and colleagues, a likely difference that may contribute to our divergent findings is the complexity of the coding task. While many classification tasks may only require a couple of categories and likely yield results similar to Törnberg’s, future appliers of the GPT framework in political science should be mindful of the likely loss of accuracy associated with complex multiclass classification for GPT. However, as the GPT framework has been consistently updated, and will likely continue to be further improved upon, future research should assess upcoming versions of GPT as its accuracy at any given stage will be an empirical question.

While comparing our results to other studies of the GPT framework helps situate our findings in the emerging literature on the use of GPT, it is also necessary to situate our findings relative to other high-profile examples of classification in political science to gage the accuracy of GPT relative to standards within the field. This may yield insights valuable use-cases for GPT coding. The case that may be most directly comparable to ours is Loftis \& Mortensen (2020) who use the CAP scheme and naive bayes classification on Danish municipal council meetings and achieve accuracy scores around 67 to 75. As the accuracy scores of our GPT test with minimum human intervention fall within 67 to \%69 GPT may be a viable alternative to Naive Bayes classification without a substantial accuracy loss and without a substantive increase in human labor.

One possible caveat of using GPT for classification is apparent when we consider precision results within different categories. While the best performing categories achieve as high a precision as 0.86-0.89 a few others achieved 0.27 to 0.32 (see appendix B, table 5-10). What is furthermore of note is that the topic-accuracy varies between GPT 3.5 and GPT 4 so that it is not necessarily the same topics that are accurately categorized, suggesting that further development of the algorithm is needed before it can be reliably used for classification tasks for specific topics. The most problematic case is the immigration topic, when coded according to scenario 1, which achieves an unsatisfactory low 0.28 using the GPT 4.0, but had shown very high performance using GPT 3.5 resulting in a 0.85 accuracy score. However, an overall comparison of within topic accuracy scores between the three scenarios suggests a vast improvement when scenario 3 is considered, thus suggesting that this may be the best overall strategy to decrease variation in accuracy scores across topics. Further studies should seek to evaluate within-topic accuracy scores to see if a pattern emerges in which topics are consistently underperforming and may therefore need extra manual quality control on the backend of the process.

While our results suggest that the GPT models do not present with sufficient accuracy compared to what is currently achieved in the literature using problem-tailored algorithms, there may still be easily overlooked advantages that in the future could tip the scale towards large language models like GPT. Most of the literature for which comparable accuracy estimates can be obtained presents in-sample test accuracy results, e.g. testing the results of the training set against a subset of the data achieved by splitting the sample in training and test set. Terechshenko and colleagues (2020) present a study with highly comparable parameters to ours, which tests both in and out of sample prediction of a variety of categorization algorithms. Furthermore, they also rely on congressional bills and the CAP scheme. Comparing our results, they systematically find that in-sample predictions which outperforms the accuracy achieved from our GPT whether the model fitted is based on linear SVM, Logistic Regression, Random Forest, Roberta, Ulmfit or XLNET. However, the out-of-sample prediction rates, as tested by using their algorithms trained on CAP hearings on New York Times CAP data, fail to rival the accuracy level of our tests with the GPT models for all of their specifications. Given that coding using GPT essentially could be thought of as out-of-sample prediction from the get-go, this comparison highlights the exact advantage of the GPT framework – it has fairly high accuracy across different contexts. As the ultimate goal of training models in political science is usually out-of-sample prediction, this could highly favor GPT as a coding tool.

However, whether our test is truly out-of-sample prediction rests on one crucial assumption that future users of GPT in political science must be consciously aware of. One of the current issues with relying on algorithms pretrained by external actors is that there is limited transparency in where the sources of training data are collected from.  On the one hand, seeing as many data sources in political science are publicly available, they may already have been included in the model’s training and may therefore boost the model’s ability to classify these particular texts. On the other hand, it might mean that our accuracy in these tests is biased either up or down dependent on whether the data has been used in the pre-training of GPT. Following from this logic, the ability to generalize our results to other use cases, that may themselves be build on data that could possibly have been used in training GPT may sometimes be limited. However, such generalization must be assessed on a case-by-case comparison of the data future users want to explore using GPT. As a possible baseline, we consider the public availability of congressional bills and hearings fairly likely to have been included in the training, but there is truly no way of knowing without increased transparency from OpenAI.

In line with other studies, our results also suggest that prompt engineering is an important aspect of using GPT to perform classification tasks. Importantly, however, our results also suggest that GPT 4 is much less susceptible performance enhancement as a result of prompt engineering than GPT 3.5. In turn, this indicates that while future uses of GPT for classification tasks need to consider prompt engineering as a mandatory aspect of GPT work, the importance is likely to decrease with future updates.

 GPT models, except the GPT4, have now been made available for fine-tuning. This will mean that with relatively little additional effort the algorithm may be developed with further training to overcome possible issues with field specificity. It may be the case, that classification of political documents achieve a relatively high base line of accuracy given that many political text sources are freely available through the internet and may have been part of the data the GPT model was trained on. This may be less likely for fringe issues that are politically less salient or maybe country specific, and accuracy may suffer as a result when using a GPT framework on edge cases, rare political subjects, or data from eras with less publicly available digitized data. Furthermore, while this study’s results show the eventual promise of the GPT framework in coding tasks on political texts, future research will have to determine how well this extends beyond the political sphere. Caution is advised in using GPT on other types of texts before performance has been evaluated.

\section*{Conclusion}

In this paper we examined the viability of two versions of the GPT model for coding of policy documents. We proposed three approaches researchers might use with varying degrees of human-computer-interaction. Our results suggest that the approach based on the highest degree of human-computer-interaction allowed for an accuracy rivaling those seen in many state-of-the-art approaches while substantially cutting down the need for manual coding and therefore costs associated with classification. Mainly relying on the GPT models without substantial human intervention is currently not viable while achieving a sufficient performance, but LLMs may improve sufficiently in the future to allow for this approach to be viable. The development of open source LLM frameworks may help alleviate concerns such as black box problems relating to data input, costs, and transparency.

\section*{References}

Adler, E. S., and Wilkerson, J. (2011). The Congressional bills project.\\ http://www.congressionalbills.org.\\
\newline
Argyle, L., Busby, E., Fulda, N., Gubler, J., Rytting, C., \& Wingate, D. (2023). Out of One, Many: Using Language Models to Simulate Human Samples. Political Analysis, 31(3), 337-351. doi:10.1017/pan.2023.2 \\
\newline
Brown, T.B.; Mann, B; Ryder,; Subbiah, M; Kaplan, J; Dhariwal, P; Neelakantan, A;  Shyam, P; Sastry, Askell, A.G. et al. Language models are few-shot learners. In NeurIPS, 2020. https://doi.org/10.48550/arXiv.2005.14165 \\
\newline
Burscher, B; Vliegenthart, R \& De Vreese, C. H. (2015). Using Supervised Machine
Learning to Code Policy Issues: Can Classifiers Generalize across Contexts? The ANNALS
of the American Academy of Political and Social Science, 659(1), 122–131.\\
\newline 
Caramancion, K. M. (2023). News Verifiers Showdown: A Comparative Performance Evaluation of ChatGPT 3.5, ChatGPT 4.0, Bing AI, and Bard in News Fact-Checking. arXiv:2306.17176. [cs.CL]. \href{https://doi.org/10.48550/arXiv.2306.17176}{https://doi.org/10.48550/arXiv.2306.17176}\\
\newline
Eshima, Imai, K., \& Sasaki, T. (2023). Keyword-Assisted Topic Models. American Journal of Political Science. https://doi.org/10.1111/ajps.12779\\
\newline
Gentzkow, Kelly, B., \& Taddy, M. (2019). Text as Data. Journal of Economic Literature, 57(3), 535–574. https://doi.org/10.1257/jel.20181020\\
\newline
Gilardi, Alizadeh, M., \& Kubli, M. (2023). ChatGPT outperforms crowd workers for text-annotation tasks. Proceedings of the National Academy of Sciences - PNAS, 120(30), \\ e2305016120–e2305016120. https://doi.org/10.1073/pnas.2305016120 \\
\newline
Hearings. The Policy Agendas Project at the University of Texas at Austin, (2017).\\ www.comparativeagendas.net. Accessed February 28, 2023\\
\newline
Kocoń, Cichecki, I., Kaszyca, O., Kochanek, M., Szydło, D., Baran, J., Bielaniewicz, J., Gruza, M., Janz, A., Kanclerz, K., Kocoń, A., Koptyra, B., Mieleszczenko-Kowszewicz, W., Miłkowski, P., Oleksy, M., Piasecki, M., Radliński, Ł., Wojtasik, K., Woźniak, S., \& Kazienko, P. (2023). ChatGPT: Jack of all trades, master of none. Information Fusion, 99, 101861–. https://doi.org/10.1016/j.inffus.2023.101861 \\
\newline
Lazer, Pentland, A., Adamic, L., Aral, S., Barabási, A.-L., Brewer, D., Christakis, N., Contractor, N., Fowler, J., Gutmann, M., Jebara, T., King, G., Macy, M., Roy, D., \& Van Alstyne, M. (2009). Social science: Computational social science. Science (American Association for the Advancement of Science), 323(5915), 721–723. https://doi.org/10.1126/science.1167742\\
\newline
Loftis, M., \& Mortensen, P. B. (2020). Collaborating with the Machines: A hybrid method for
classifying policy documents. Policy Studies Journal, 48(1), 184-206. \\ https://doi.org/10.1111/psj.12245 \\
\newline
Sebők, M., \& Kacsuk, Z. (2020). The Multiclass Classification of Newspaper Articles with Machine Learning: The Hybrid Binary Snowball Approach. Political Analysis, 29, 236–249. https://doi.org/10.1017/pan.2020.27\\
\newline 
Eshima, Imai, K., \& Sasaki, T. (2023). Keyword-Assisted Topic Models. American Journal of Political Science. https://doi.org/10.1111/ajps.12779\\
\newline
Törnberg, P. (2023). ChatGPT-4 Outperforms Experts and Crowd Workers in Annotating Political Twitter Messages with Zero-Shot Learning. arXiv:2304.06588. [cs.CL].\\ \href{https://doi.org/10.48550/arXiv.2304.06588}{https://doi.org/10.48550/arXiv.2304.06588} \\
\newline
Wu, P. Y., Nagler, J., Tucker, J. A., \& Messing, S. (2023). Large Language Models Can Be Used to Estimate the Latent Positions of Politicians. arXiv:2303.12057. [cs.CY]. \\https://doi.org/10.48550/arXiv.2303.12057\\
\newline
Zhu, Y., Zhang, P., Haq, E.-U., Hui, P., \& Tyson, G. (2023). Can ChatGPT Reproduce Human-Generated Labels? A Study of Social Computing Tasks. arXiv:2304.10145. [cs.AI]. \href{https://doi.org/10.48550/arXiv.2304.10145}{https://doi.org/10.48550/arXiv.2304.10145} \\
\newline
\vskip 40pt
\newpage

\begin{appendices}
\section{Prompt Examples}

\subsection{GPT 3.5 Prompt Example}

Here is a dictionary which contains policy issue labels (keys) and specific topics (values) associated with those labels : \\
''' \{'Macroeconomics': Includes issues related to general domestic macroeconomic policy and more specific issues including inflation, cost of living, prices, interest rates, unemployment rate, impact of unemployment, training and retraining, unemployment benefits, monetary policy, central bank, treasury, public debt, budgeting, efforts to reduce deficits, tax policy, the impact of taxes, and tax enforcement, manufacturing policy, industrial revitalization and growth, wage or price control, emergency price controls.,
\\'Civil Rights': Includes issues related to civil rights and minority rights. More specifically, it includes issues related to minority, ethnic, and racial group discrimination, sex, gender, and sexual discrimination, age discrimination, mandatory retirement age policies, handicap and disease discrimination, voting rights, expanding or contracting the franchise, participation and related issues, freedom of speech, religious freedoms, other types of freedom of expression, privacy rights, privacy of records, access to government information, abortion rights and anti-government activity groups.,
\\'Health': Includes issues related generally to health care, including appropriations for general health care government agencies. More specifically, it includes issues related to broad, comprehensive changes in the health care system, health insurance reform, regulation, availability, and cost of insurance, the regulation and promotion of pharmaceuticals, medical devices, and clinical labs, facilities construction, regulation and payments, provider and insurer payments and regulation, medical liability, malpractice issues, medical fraud and abuse, and unfair practices, the supply and quantity of labor in the health care industry, training and licensing, disease prevention and treatment, health promotion, coverage and quality of infants and children care, school health programs, mental health care, and mental health disease, long term care, home health care, the terminally ill and rehabilitation services, prescription drug coverage, programs to pay for prescription drugs, policy to reduce the cost of prescription drugs, tobacco abuse, treatment, education and health effects, alcohol and illegal drug abuse, treatment, education and health effects, health care research and development.‚
\\'Agriculture': Agriculture policy, the regulation and impact of agricultural foreign trade, government subsidies to farmers and ranchers, food inspection and safety, the regulation of agricultural marketing and providing information to consumers regarding a healthy diet, animal and crop disease, pest control and pesticide regulation, and welfare for domesticated animals, fishing, commercial fishery regulation and conservation, agricultural research and development., 
\\'Labor': Employment, and pensions, worker safety and protection and compensation for work-related injury and disease, job training for adult workers, workforce development, and efforts to retrain displaced workers, employee benefits, pensions, and retirement accounts, including government-provided unemployment insurance, labor unions, collective bargaining, employer-employee relations, fair labor standards, minimum wage and overtime compensation, labor law, youth employment, child labor and job training for youths, migrant, guest and seasonal workers.,
\\'Education': Education policy, higher education, student loans and education finance, the regulation of colleges and universities, elementary and private schools, school reform, safety in schools, efforts to generally improve educations standards and outcomes, education of underprivileged students, adult literacy programs, bilingual education needs, rural education initiatives, vocational education for children and adults, special education for the physically or mentally handicapped, education excellence, research and development in education.,
\\'Environment': Environmental policy, domestic drinking water safety, supply, pollution and additives, disposal and treatment of wastewater, solid water and runoff, hazardous waste and toxic chemical regulation, treatment and disposal, air pollution, climate change, noise pollution, recycling, reuse, resource conservation, environmental hazards, indoor air contaminations, indoor hazardous substances, forest protection, endangered species, control of the domestic illicit trade in wildlife products, regulation of laboratory or performance animals, land and water resource conservations, research and development in environmental technology, not including alternative energy.,
\\'Energy': Energy policy, nuclear energy, safety and security, and disposal of nuclear waste, general electricity, hydropower, regulation of electric utilities, natural gas and oil, drilling, oil spills and flaring, oil and gas prices, shortages and gasoline regulation, coal production, use, trade, and regulation, alternative and renewable energy, biofuels, hydrogen and geothermal power, energy conservation and energy efficiency, energy research and development.‚
\\'Immigration': Immigration, refugees, and citizenship.,
\\'Transportation': Transportation policy, mass transportation construction, regulation, safety and availability, highway construction, maintenance and safety, air travel, regulation and safety of aviation, airports, air traffic control, pilot training, aviation technology, railroads, rail travel, rail freight, development and deployment of new rail technologies, maritime transportation, inland waterways and channels, infrastructure and public works, transportation research and development.,
\\'Law and Crime': General law, crime and family issues, law enforcement agencies, white collar crime, organized crime, counterfeiting and fraud, cyber-crime, money laundering, illegal drug crime, criminal penalties for drug crimes, court administration, judiciary appropriations, guidelines for bail, pre-release, fines and legal representation, prisons and jails, parole systems, juvenile crime and justice, juvenile prisons and jails, efforts to reduce juvenile crime and recidivism, domestic violence, child welfare, family law, domestic criminal and civil codes, control, prevention and impact of crime, police and other general domestic security responses to terrorism.‚
\\'Social Welfare': Social welfare policy, poverty assistance, food assistance programs, programs to assess or alleviate welfare dependency and tax credits directed at low-income families, elderly issues and elderly assistance, government pensions, aid for people with physical or mental disabilities, domestic volunteer associations, charities and youth organizations, parental leave and childcare.,
\\'Housing': Housing and urban affairs, housing and community development, neighborhood development, national housing policy, urban development and general urban issues, rural housing, economic, infrastructure and other development in non-urban areas, housing for low-income individuals and families, housing for the elderly, housing for homeless, efforts to reduce homelessness, housing and community development research and development.‚
\\'Domestic Commerce': Government agencies regulating domestic commerce, the regulation of national banking systems and other non-bank financial institutions, the regulation and facilitation of securities and commodities trading, regulation of investments and related industries and exchanges, consumer finance, mortgages, credit cards, access to credit records, consumer credit fraud, insurance regulation, fraud and abuse in the insurance industry, the financial health of insurance industry, insurance availability and cost, personal, commercial and municipal bankruptcies, corporate mergers, antitrust regulation, corporate accounting and governance, corporate management, small businesses, copyrights and patents, patent reform, intellectual property, domestic natural disaster relief, disaster or flood insurance, natural disaster preparedness, tourism regulation, promotion and impact, consumer fraud and safety in domestic commerce, regulation and promotion of sports, gambling and personal fitness, domestic commerce research and development.‚
\\'Defense': General defense policy, defense alliance and agreement, security assistance, UN peacekeeping activities, military intelligence, espionage and covert operations, military readiness, coordination of armed services air support and sealift capabilities, national stockpiles of strategic materials, nuclear weapons, nuclear proliferation, modernizations of nuclear equipment, military aid to other countries, control of arms sales, military manpower, military personnel and their dependents, military courts, general veterans‚Äô issues, military procurement, conversion of old equipment, weapons and systems evaluation, military installations, construction and land transfers, military reserves and reserve affairs, military nuclear and hazardous waste disposal and military environmental compliance, domestic civil defense, national security responses to terrorism, non-contractor civilian personnel, civilian employment in the defense industry, military base closings, military contractors, war-related military operations, prisoners of war and collateral damage to civilian populations, claims against the military, settlements for military dependents, compensation for civilians injured in military operations, defense research and development.‚
\\'Technology': Space, science, technology and communications, the government use of space and space resource exploitation agreements, government space programs and space exploration, military use of space, regulation and promotion of commercial use of space, commercial satellite technology, government efforts to encourage commercial space development, science and technology transfer, international science cooperation, telephones and telecommunication regulation, infrastructure for high speed internet, mobile phones, broadcast industry, published media, control of the electromagnetic spectrum, weather forecasting, oceanography, geological surveys, weather forecasting research and technology, computer industry, regulation of the internet, cyber security.‚
\\'Foreign Trade': Negotiations, disputes and agreements, including tax treaties, export regulation, subsidies, promotion and control, international private business investment and corporate development, productivity of competitiveness of domestic businesses and balance of payment issues, tariffs and other barriers to imports, import regulation and impact of imports on domestic industries, exchange rate and related issues.,
\\'International Affairs': General international affairs and foreign aid, foreign aid not directly targeting and increasing international development, international resources exploitations and resources agreements, law of the sea and international ocean conservation efforts, developing countries, international finance and economic development banks, sovereign debt and implications for international lending institutions, Western Europe and European Union, foreign country or region, assessment of political issues in other countries, relations between individual countries, human rights violations, human rights treaties and conventions, UN reports on human rights, crimes associated with genocide or crimes against humanity, international organizations, NGOs, the United Nations, International Red Cross, UNESCO, International Olympic Committee, International Criminal Court, international terrorism, hijacking, acts of piracy in other countries, efforts to fight international terrorism, international legal mechanisms to combat terrorism, diplomats, diplomacy, embassies, citizens abroad, foreign diplomats in the country, visas and passports.
\\'Government Operations': Government agencies, intergovernmental relations, local government issues, general government efficiencies, bureaucratic oversight, postal services, regulation of mail, post-civil service, government pensions, general civil service issues, nominations and appointments, currency, national mints, medals and commemorative coins, government procurement, government contractors, contractor and procurement fraud, procurement processes and systems, government property management, construction and regulation, tax administration, enforcement, auditing for both individuals and corporations, public scandal and impeachment, government branch relations, administrative issues, constitutional reforms, regulation of political campaigns, campaign finance, political advertising and voter registration, census and statistics collection by government, capital city, claims against the government, compensation for the victims of terrorist attacks, compensation policies without other substantive provisions, national holidays and their observation.,
\\'Public Lands': General public lands, water management, and territorial issues, national parks, memorials, historic sites and recreation, the management and staffing of cultural sites, indigenous affairs, indigenous lands, assistance to indigenous people, natural resources, public lands, forest management, forest fires, livestock grazing, water resources, water resource development and civil works, flood control and research, territorial and dependency issues and devolution.‚
\\'Culture': General cultural policy issues.
\}

                            Here is the title of a congressional bill: 
 
                            """1: A bill to amend the Foreign Assistance Act of 1961 to require congressional authorization for the involvement of American Forces in further hostilities in Indochina, and for extending assistance to North Vietnam.
 """
 \\
 Map it to one of the policy issue labels listed in the dictionary at the top. Use the specific topics information contained in the dictionary to infer the correct label. If the title is not related to any of the labels in the policy issue categories dictionary, then assign "Other" label. Give me only the predicted policy issue label. No explanation needed. Do not assign a label which is not one of the keys of the dictionary or the label 'Other'. The generated response must only include the predicted label, nothing else.

\subsection{GPT4 Prompt Example}

Here is a list of policy issue topic labels:
"""
\\
Macroeconomics, \\
Civil Rights, \\
Health, \\
Agriculture, \\
Labor, \\
Education, \\
Environment, \\
Energy, \\
Immigration, \\
Transportation, \\
Law and Crime, \\
Social Welfare, \\
Housing, \\
Domestic Commerce, \\
Defense, \\
Technology, \\
Foreign Trade, \\
International Affairs, \\
Government Operations, \\
Public Lands, \\
Culture, \\
Other. \\
""" \\
Here are some sentences: 

"""\\
1: To amend title II of the Social Security Act so as to liberalize the conditions governing eligibility of blind persons to receive disability insurance benefits .thereunder
2: A bill to amend title V of the Elementary and Secondary Education Act of 1965 to reduce class size through the use of highly qualified teachers, and for other purposes.
3: To establish an emergency community facilities arid public works program in the Community Facilities Administration of the Housing and Home Finance Agency
4: To direct the Secretary of Health and Human Services to make payments to each State for the operation of a comprehensive health insurance plan ensuring health insurance coverage for individuals and families in the State, and for other purposes.
5: For the relief of Isolene E. F. Shakespeare
6: A bill to extend temporarily the suspension of duty on mucochloric acid.
7: A bill to amend the Fair Packaging and Labeling Act to require certain labeling to assist the consumer in purchases of packaged perishable or semiperishable foods.
8: A bill to authorize appropriations to the National Aeronautics and Space Administration for research and development, space flight, control and data communications, construction of facilities, and research and program management, and for other purposes.
9: A bill to provide for the resolution of claims and disputes relating to the Government contracts awarded by executive agencies.
10: A bill to revise, reform, and standardize the organization of the executive part of the military departments, and for other purposes.
11: A bill to expand the middle class, reduce the gap between the rich and the poor, keep our promises to veterans, lower the poverty rate, and reduce the Federal deficit by repealing tax breaks for the wealthiest one percent and eliminating unnecessary Cold War era defense spending, and for other purposes.
12: A bill to institute withholding tax on dispositions of United States real property interests by non-United States persons and to repeal present reporting requirements in such transactions.
13: For the relief of Teofila Pardo Ruiz
14: A bill to require a Federal permit for the taking of any migratory game birds other than migratory waterfowl.
15: To provide for orderly trade in textile articles and articles of leather footwear, and for other purposes
16: For the relief of Joshua Felise Ziro Brevio
17: To amend title XVIII of the Social Security Act to enhance beneficiary access in rural areas to quality health care services under the Medicare Program.
18: A bill for the relief of Bhuminder Singh
19: To create a catalog of Federal assistance programs, and for other purposes
20: For the relief of Mr. Pasquale D'Addio
21: To control unfair trade practices affecting producers of agricultural products and associations of such producers, and for other purposes
22: To revise the Federal laws, and for other purposes
23: A bill to amend title 10, United States Code, to provide greater flexibility for the Armed Forces in ordering Reserves to active duty, and for other purposes.
24: For the relief of Angero and Theophilos Kamperldes
25: To amend title 5, United States Code, to provide a pay differential of 10 percent for night work performed by employees subject to prevailing rate pay systems, and for other purposes
26: A bill for the relief of Dr. Takashi Sawa.
27: To amend title 18, United States Code, to provide the penalty of death for certain murders of State and local correctional officers by incarcerated persons, and for other purposes.
28: To amend title 38 of the United States Code to permit members of the Department of Medicine and Surgery of the Veterans Administration to be granted leave to pursue certain courses of study
29: To authorize and request the President to award a Presidential Unit Citation to the 761st Tank Battalion
30: A bill to authorize appropriations for the preservation and restoration of historic buildings at historically black colleges and universities.
31: To authorize the extension of a loan of a naval vessel to the Government of the Republic of China
32: To establish a U.S. Disarmament Agency for World Peace and Security
33: To amend the Communications Act of 1934 to require Internet access providers to provide screening software to permit parents to control Internet access by their children.
34: To extend the existing suspension of duty on 4-Chloro-3-methylphenol.
35: To amend the privacy provisions of title 5, United States Code, to improve the protection of individuals information and to reestablish a permanent Privacy Protection Commission as an independent entity in the Federal Government, and for other purposes.
36: For .the relief of Jose Carrasco Hendy
37: A bill to amend title XVIII of the Social Security Act to provide for the administrative and judicial review of claims (involving the amount of benefits payable) which arises under the supplementary medical insurance program.
38: A bill to extend to all unmarried individuals the full tax benefits of income splitting now enjoyed by married individuals filing joint returns.
39: A bill to amend the Small Business Act to provide assistance to areas of economic dislocation in the United States that is the result of drastic fluctuation in the value of the currency in a country contiguous to the United States.
40: To amend the Immigration and Nationality Act, and for other purposes
41: To prohibit the exportation of Alaskan North Slope crude oil.
42: For the relief of Florencia H. Fernandez
43: A bill to authorize and facilitate the improvement of water management by the Bureau of Reclamation, to require the Secretary of the Interior and the Secretary of Energy to increase the acquisition and analysis of water resources for irrigation, hydroelectric power, municipal, and environmental uses, and for other purposes.
44: For the relief of Melbourne Murray
45: To repeal section 115 of the Clean Air Act.
46: A bill to provide for collection by the Federal Government of data regarding service sector economic activity.
47: To increase from 8600 to \$1,000 the personal income tax exemptions of a taxpayer (including the exemption for a spouse, the exemption for a dependent, and the additional exemptions for old age or blindness)
48: A bill to repeal the recently enacted provisions authorizing increases in the salaries of Senators and Representatives.
49: To improve the reading and literacy skills of children and families by improving in-service instructional practices for teachers who teach reading, to stimulate the development of more high-quality family literacy programs, to support extended learning-time opportunities for children, to ensure that children can read well and independently not later than third grade, and for other purposes.
50: A bill to amend the Internal Revenue Code of 1986 to clarify the treatment of incentive stock options and employee stock purchase plans.
51: To amend title 5, United States Code, to correct certain inequities in the crediting of National Guard technician service in connection with civil service retirement, and for other purposes
52: A bill to provide for permanent tax reductions for individuals and businesses.
53: A bill for the relief of Joseph Veich, also known as Guiseppe Veic
54: To amend the Federal Credit Union Act
55: A bill to provide for the temporary suspension of duty on the importation of fluorspar.
56: To establish a commission to study employment and economic insecurity in the workforce in the United States.
57: A bill to establish an interagency committee and an advisory council to examine and investigate the causes and effects of atmospheric fallout and to develop a comprehensive plan to reduce such fallout.
58: A bill for the relief of Mary Eliadou Rutherford
59: To amend section 406 (b) of the Civil Aeronautics Act of 1938, so as to eliminate authority for the payment of subsidies for domestic trunk air transportation
60: To amend title III of the Older Americans Act of 1965 with respect to assistance to older individuals who reside in rural areas.
61: A bill to provide for a temporary increase in the public debt limit.
62: To amend title 23 of the United States Code to authorize parts of the National System of Interstate and Defense Highways to be constructed underground, to provide adequate public shelters in case of nuclear and germ warfare, and for other purposes
63: To amend the Illegal Immigration Reform and Immigrant Responsibility Act of 1996 to require the Immigration and Naturalization Service to verify whether an alien has an immigration status rendering the alien eligible for service in the Armed Forces of the United States and to achieve parity between the immigration status required for employment as an airport security screener and the immigration status required for service in the Armed Forces, and to amend the Immigration and Nationality Act to permit naturalization through active-duty military service during specified military operations.
64: A bill to amend the Foreign Service Act of 1980, title 5 of the United States Code, and the Hostage Relief Act of 1980, and for other purposes.
65: For the relief of Mrs. Vassiliki P. Theodorou
66: A bill to amend the Federal Trade Commission Act (15 U.S.C. 41) to provide that under certain circumstances exclusive territorial arrangements shall not be deemed unlawful.
67: To establish a cropland retirement program
68: For the relief of Philip I. Veira
69: To provide a 5-year program of assistance to enable depressed segments of the fishing industry in the United States to regain a favorable economic status, and for other purposes
70: To amend the Internal Revenue Code of 1986 to provide a credit against tax for energy efficient appliances.
71: To amend the Federal Trade Commission Act to provide for the issuance of temporary cease-and-desist orders to prevent certain acts and practices pending completion of Federal Trade Commission proceedings
72: To provide means of further securing and protecting the right to vote.
73: To amend the Mineral Leasing Act for Acquired Lands (61 Stat. 913) to provide for competitive bidding on oil and gas leases issued thereunder where a competitive interest in leasing is evident
74: A bill to provide for the disposition of funds appropriated to pay a judgment in favor of the Quileute Tribe of Indians, including the Hoh Tribe, and for other purposes
75: To amend section 2(2) of the National Labor Relations Act to provide that the definition of employer therein shall include certain corporations or associations operating a hospital
76: To improve patient access to health care services and provide improved medical care by reducing the excessive burden the liability system places on the health care delivery system.
77: To amend title 38 of the United States Code so as to extend the period of entitlement of persons entitled to assistance under the veterans educational assistance program and the war orphans educational assistance program, and for other purposes
78: A bill to amend the District of Columbia Self-Government and Government Reorganization Act to repeal the limitation on the council of the District of Columbia regarding the imposition of any tax upon individuals who are not residents of the District of Columbia.
79: To provide for the joint appointment of the Architect of the Capitol by the Speaker of the House of Representatives, the Majority Leader of the Senate, the Minority Leaders of the House of Representatives and Senate, and the chairs and ranking minority members of the committees of Congress with jurisdiction over the Office of the Architect of the Capitol, and for other purposes.
80: To amend title 13, United States Code, to limit the categories of questions required to be answered under penalty of law in the decennial censuses of population, unemployment, and housing, and for other purposes
81: A bill to establish demonstration projects to provide at-home infant care benefits.
82: For the relief of Eufemio Dacanay Felix
83: To provide for the suspension of U.S. foreign aid to any country with respect to which diplomatic relations with the United States have been terminated or suspended
84: To direct the Administrator of the National Oceanic and Atmospheric Administration to establish an integrated Federal ocean and coastal mapping plan for the Great Lakes and coastal state waters, the territorial sea, the exclusive economic zone, and the Continental Shelf of the United States, and for other purposes.
85: A bill for the relief of Isoo Kojima.
86: To amend the Outer Continental Shelf Lands Act to require the chief executive officer of each drilling and production operation under a lease under that Act to annually certify the operator's compliance with all applicable laws and operating regulations.
87: For the relief of Onofrio Saia
88: To make certain expenditures by the city of Sanford, N.C., eligible as local grants-in-aid for purposes of title I of the Housing Act of 1949
89: A bill for the relief of Major Joseph R. Williams.
90: A bill to authorize the establishment of postal stations and branch post offices at camps, posts, or stations of the Armed Forces (including the Coast Guard), and at defense or other strategic installations, and for other purposes
91: A bill entitled the Federal Election Commission Authorization.
92: A bill to suspend temporarily the duty on Capa 4000-series.
93: A bill to preserve, protect, and maintain the original boundary stones of the Nation's Capital.
94: To amend chapter 1of the Internal Revenue Code of 1954, in regard to certain investments by dealers in real property
95: To suspend until January 1, 1995, the duty on Malathion.
96: To amend the Internal Revenue Code of 1954 to provide for the determent of income from service contracts
97: A bill to amend part B of title IV of the Social Security Act to provide, as the primary form in which services are to be furnished under the child-welfare services program, for supportive day treatment and in-home services to children and families.
98: A bill for the relief of Chunchai Letasz.
99: To amend section 8(c) of theFederal Water Pollution Control Act toprovide a new formula for the allocation of construction grant funds for the fiscal years beginning on or after July 1, 1970
100: For the relief of BienvenidoTan\\
Please assign a policy issue topic label to each sentence. Give me only the 
sentence number and the label, and put one label per line. Don't use any label which is not listed in the 
policy issue topics array. Use the "Other" label if the sentence is not related to any of the first 21 
categories with a very high probability.\\

Here are two example bill titles that belongs to the other category: \\
                1: 'A bill for the relief of Andrew D, Sumner.'\\
                2: 'A bill for the relief of Felix Hernandez-Arana and his wife, Felicia Ogaldez-De Hernandez.\\
Here are some notes: \\
                - The category defense also includes issues related to veterans affairs and welfare benefits to current military members, \\
                - The category public lands encompasses issues including, but not limited to, Native Americans affairs, national parks and national forests and interstate highways, \\
                - Abortion related issues belong to the category of civil rights, \\
                - Banking and finance related issues are related to domestic commerce category,
                - Issues related countries other than the United States are related to the category of international affairs, \\
                - Technology category also includes issues related to science and space, \\
                - Titles which contain keywords such as import, export, tariff and duty are related to the category of foreign trade, \\
                - Issues related to American state governments and the federal government organizations and their employees, not including the members of the army, are related to government operations category, \\
                - Culture category is very rare, less than one percent, in the congressional bill title data. \\ 
                Here is an example title that belongs to the government operations category: \\
                'To retain coverage under the laws providing employee benefits, such as compensation for injury, retirement, 
                life insurance, and health benefits for employees of the Government of the United States who transfer to 
                Indian tribal organizations to perform services in connection with governmental or other activities 
                which are or have been performed by Government employees in or for Indian communities, and for other purposes' \\

            You must give me a list with 100 labels.

\section{Additional Results}

\begin{figure}[htp]
\centering
\includegraphics[width=10cm]{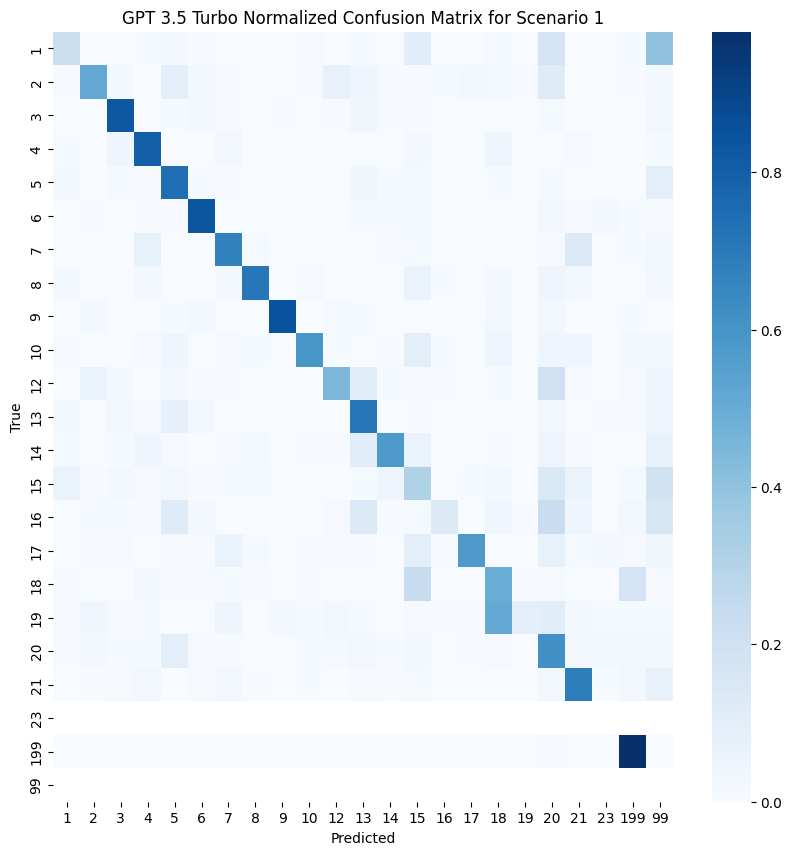}
\caption{}
\label{fig:image}
\end{figure}

\begin{figure}[htp]
\centering
\includegraphics[width=10cm]{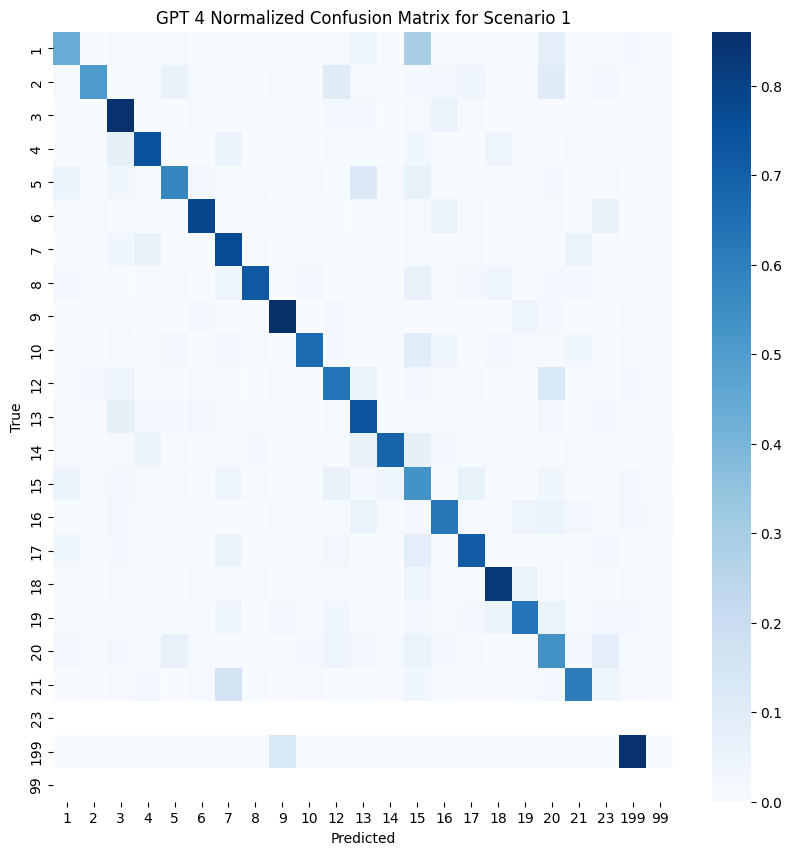}
\caption{}
\label{fig:image}
\end{figure}

\begin{figure}[htp]
\centering
\includegraphics[width=10cm]{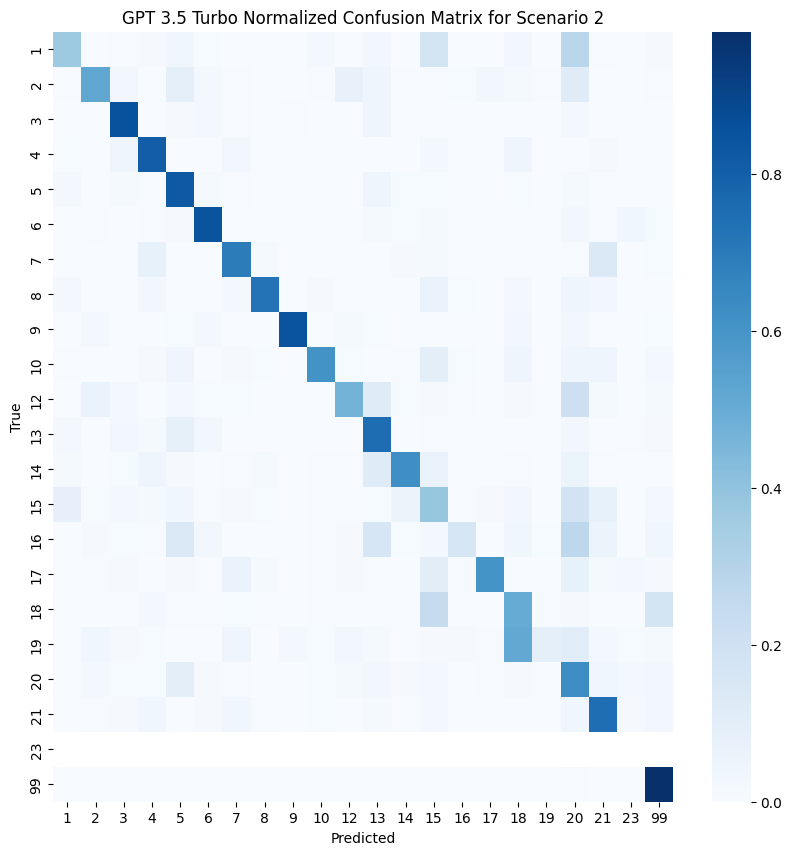}
\caption{}
\label{fig:image}
\end{figure}

\begin{figure}[htp]
\centering
\includegraphics[width=10cm]{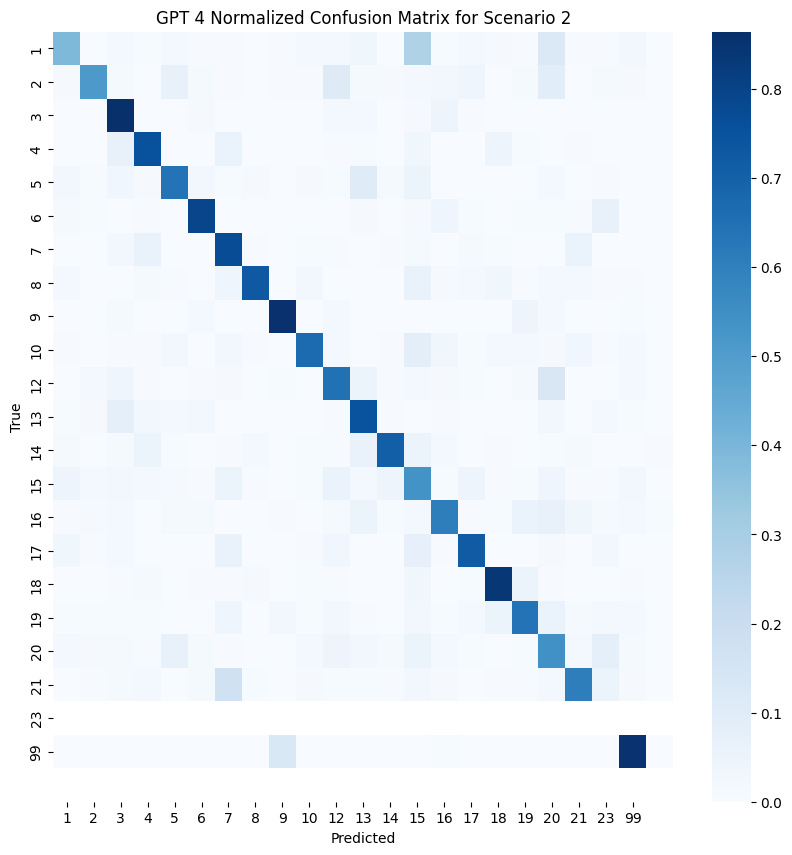}
\caption{}
\label{fig:image}
\end{figure}

\begin{figure}[htp]
\centering
\includegraphics[width=10cm]{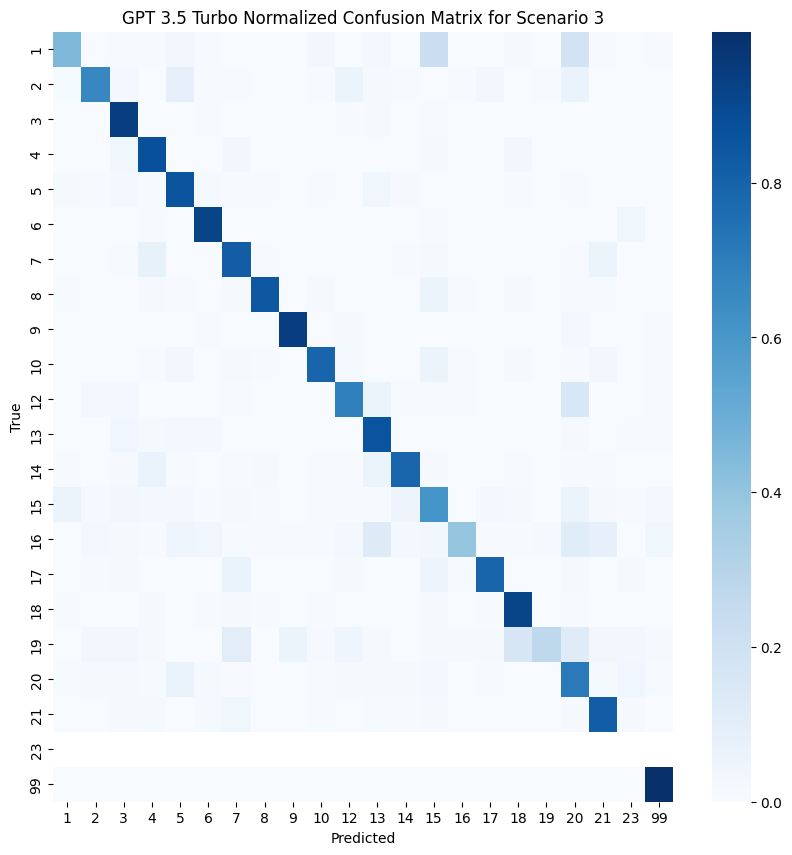}
\caption{}
\label{fig:image}
\end{figure}

\begin{figure}[htp]
\centering
\includegraphics[width=10cm]{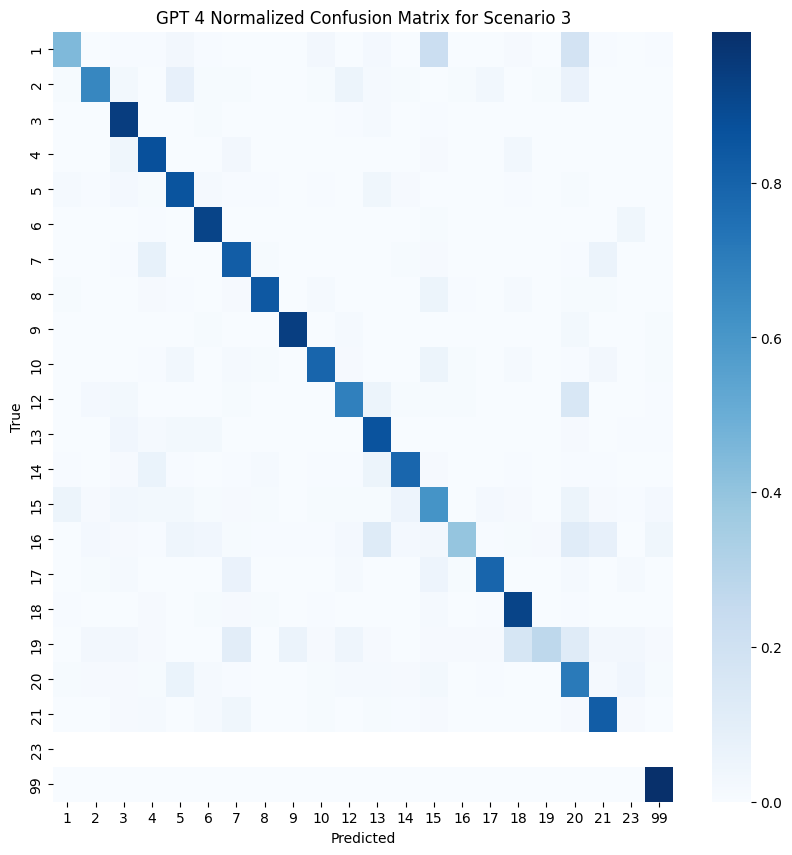}
\caption{}
\label{fig:image}
\end{figure}

\begin{table}[h]
\centering
\caption{Class-specific performance metrics for GPT 3.5 Turbo in Scenario 1 - Congressional Bills}
\begin{tabular}{|c|c|c|c|}
\hline
Class & Precision & Recall & F1-score \\
\hline
Macroeconomics & 0.49 & 0.22 & 0.31 \\
Civil Rights & 0.49 & 0.52 & 0.50 \\
Health & 0.85 & 0.83 & 0.84 \\
Agriculture & 0.70 & 0.80 & 0.74 \\
Labor & 0.43 & 0.74 & 0.55 \\
Education & 0.75 & 0.83 & 0.79 \\
Environment & 0.67 & 0.67 & 0.67 \\
Energy & 0.86 & 0.71 & 0.78 \\
Immigration & 0.85 & 0.85 & 0.85 \\
Transportation & 0.87 & 0.59 & 0.70 \\
Law and Crime & 0.79 & 0.45 & 0.57 \\
Social Policy & 0.47 & 0.71 & 0.56 \\
Housing & 0.63 & 0.58 & 0.60 \\
Domestic Commerce & 0.32 & 0.31 & 0.32 \\
Defense & 0.80 & 0.13 & 0.23 \\
Technology & 0.79 & 0.58 & 0.67 \\
Foreign Trade & 0.50 & 0.50 & 0.50 \\
International Affairs & 0.61 & 0.09 & 0.16 \\
Government Operations & 0.53 & 0.61 & 0.57 \\
Public Lands & 0.74 & 0.69 & 0.72 \\
Culture & 0.00 & 0.00 & 0.00 \\
Other (Private Bills) & 0.89 & 0.98 & 0.93 \\
\hline
\end{tabular}
\end{table}

\begin{table}[h]
\centering
\caption{Class-specific performance metrics for GPT 4 in Scenario 1 - Congressional Bills}
\begin{tabular}{|c|c|c|c|}
\hline
Class & Precision & Recall & F1-score \\
\hline
Macroeconomics & 0.57 & 0.44 & 0.50 \\
Civil Rights & 0.56 & 0.51 & 0.53 \\
Health & 0.75 & 0.86 & 0.80 \\
Agriculture & 0.73 & 0.75 & 0.74 \\
Labor & 0.61 & 0.58 & 0.59 \\
Education & 0.77 & 0.79 & 0.78 \\
Environment & 0.48 & 0.77 & 0.60 \\
Energy & 0.87 & 0.72 & 0.79 \\
Immigration & 0.28 & 0.86 & 0.43 \\
Transportation & 0.83 & 0.66 & 0.74 \\
Law and Crime & 0.65 & 0.64 & 0.64 \\
Social Policy & 0.56 & 0.74 & 0.64 \\
Housing & 0.68 & 0.69 & 0.68 \\
Domestic Commerce & 0.40 & 0.53 & 0.46 \\
Defense & 0.76 & 0.63 & 0.69 \\
Technology & 0.53 & 0.71 & 0.61 \\
Foreign Trade & 0.84 & 0.83 & 0.83 \\
International Affairs & 0.52 & 0.64 & 0.57 \\
Government Operations & 0.71 & 0.54 & 0.61 \\
Public Lands & 0.84 & 0.61 & 0.71 \\
Culture & 0.00 & 0.00 & 0.00 \\
Other (Private Bills) & 0.94 & 0.85 & 0.89 \\
\hline
\end{tabular}
\end{table}

\begin{table}[h]
\centering
\caption{Class-specific performance metrics for GPT 3.5 Turbo in Scenario 1 - Congressional Hearings}
\begin{tabular}{|c|c|c|c|}
\hline
Class & Precision & Recall & F1-score \\
\hline
Macroeconomics & 0.56 & 0.49 & 0.52 \\
Civil Rights & 0.54 & 0.35 & 0.43 \\
Health & 0.82 & 0.76 & 0.79 \\
Agriculture & 0.67 & 0.79 & 0.73 \\
Labor & 0.55 & 0.78 & 0.64 \\
Education & 0.74 & 0.76 & 0.75 \\
Environment & 0.71 & 0.69 & 0.70 \\
Energy & 0.80 & 0.75 & 0.78 \\
Immigration & 0.70 & 0.71 & 0.71 \\
Transportation & 0.81 & 0.66 & 0.73 \\
Law and Crime & 0.61 & 0.40 & 0.48 \\
Social Welfare & 0.37 & 0.79 & 0.50 \\
Housing & 0.66 & 0.65 & 0.65 \\
Domestic Commerce & 0.66 & 0.37 & 0.48 \\
Defense & 0.82 & 0.46 & 0.59 \\
Technology & 0.67 & 0.63 & 0.65 \\
Foreign Trade & 0.27 & 0.77 & 0.40 \\
International Affairs & 0.90 & 0.34 & 0.49 \\
Government Operations & 0.61 & 0.62 & 0.62 \\
Public Lands & 0.67 & 0.65 & 0.66 \\
Culture & 0.00 & 0.00 & 0.00 \\
\hline
\end{tabular}
\end{table}

\begin{table}[h]
\centering
\caption{Class-specific performance metrics for GPT 3.5 Turbo in Scenario 2 - Congressional Bills}
\begin{tabular}{|c|c|c|c|}
\hline
Class & Precision & Recall & F1-score \\
\hline
Macroeconomics & 0.51 & 0.24 & 0.33 \\
Civil Rights & 0.50 & 0.55 & 0.52 \\
Health & 0.85 & 0.83 & 0.84 \\
Agriculture & 0.70 & 0.80 & 0.74 \\
Labor & 0.43 & 0.74 & 0.55 \\
Education & 0.75 & 0.83 & 0.79 \\
Environment & 0.67 & 0.67 & 0.67 \\
Energy & 0.86 & 0.71 & 0.78 \\
Immigration & 0.85 & 0.85 & 0.85 \\
Transportation & 0.87 & 0.59 & 0.70 \\
Law and Crime & 0.79 & 0.45 & 0.57 \\
Social Policy & 0.47 & 0.71 & 0.56 \\
Housing & 0.63 & 0.58 & 0.60 \\
Domestic Commerce & 0.32 & 0.31 & 0.32 \\
Defense & 0.80 & 0.13 & 0.23 \\
Technology & 0.79 & 0.58 & 0.67 \\
Foreign Trade & 0.50 & 0.50 & 0.50 \\
International Affairs & 0.61 & 0.09 & 0.16 \\
Government Operations & 0.53 & 0.61 & 0.57 \\
Public Lands & 0.74 & 0.69 & 0.72 \\
Culture & 0.00 & 0.00 & 0.00 \\
Other (Private Bills) & 0.89 & 0.98 & 0.93 \\
\hline
\end{tabular}
\end{table}

\begin{table}[h]
\centering
\caption{Class-specific performance metrics for GPT 4 in Scenario 2 - Congressional Bills}
\begin{tabular}{|c|c|c|c|}
\hline
Class & Precision & Recall & F1-score \\
\hline
Macroeconomics & 0.57 & 0.44 & 0.50 \\
Civil Rights & 0.56 & 0.51 & 0.53 \\
Health & 0.75 & 0.86 & 0.80 \\
Agriculture & 0.73 & 0.75 & 0.74 \\
Labor & 0.61 & 0.58 & 0.59 \\
Education & 0.77 & 0.79 & 0.78 \\
Environment & 0.48 & 0.77 & 0.60 \\
Energy & 0.87 & 0.72 & 0.79 \\
Immigration & 0.28 & 0.86 & 0.43 \\
Transportation & 0.83 & 0.66 & 0.74 \\
Law and Crime & 0.65 & 0.64 & 0.64 \\
Social Policy & 0.56 & 0.74 & 0.64 \\
Housing & 0.68 & 0.69 & 0.68 \\
Domestic Commerce & 0.40 & 0.53 & 0.46 \\
Defense & 0.76 & 0.63 & 0.69 \\
Technology & 0.53 & 0.71 & 0.61 \\
Foreign Trade & 0.84 & 0.83 & 0.83 \\
International Affairs & 0.52 & 0.64 & 0.57 \\
Government Operations & 0.71 & 0.54 & 0.61 \\
Public Lands & 0.84 & 0.61 & 0.71 \\
Culture & 0.00 & 0.00 & 0.00 \\
Other (Private Bills) & 0.94 & 0.85 & 0.89 \\
\hline
\end{tabular}
\end{table}

\begin{table}[h]
\centering
\caption{Class-specific performance metrics for GPT 3.5 Turbo in Scenario 2 - Congressional Hearings}
\begin{tabular}{|c|c|c|c|}
\hline
Class & Precision & Recall & F1-score \\
\hline
Macroeconomics & 0.56 & 0.58 & 0.57 \\
Civil Rights & 0.54 & 0.37 & 0.44 \\
Health & 0.82 & 0.80 & 0.81 \\
Agriculture & 0.67 & 0.83 & 0.75 \\
Labor & 0.55 & 0.81 & 0.65 \\
Education & 0.74 & 0.82 & 0.78 \\
Environment & 0.71 & 0.71 & 0.71 \\
Energy & 0.80 & 0.78 & 0.79 \\
Immigration & 0.70 & 0.80 & 0.75 \\
Transportation & 0.81 & 0.75 & 0.78 \\
Law and Crime & 0.61 & 0.42 & 0.50 \\
Social Welfare & 0.37 & 0.81 & 0.50 \\
Housing & 0.66 & 0.73 & 0.69 \\
Domestic Commerce & 0.66 & 0.47 & 0.55 \\
Defense & 0.82 & 0.51 & 0.63 \\
Technology & 0.67 & 0.70 & 0.69 \\
Foreign Trade & 0.27 & 0.81 & 0.41 \\
International Affairs & 0.90 & 0.36 & 0.52 \\
Government Operations & 0.61 & 0.67 & 0.64 \\
Public Lands & 0.67 & 0.76 & 0.71 \\
Culture & 0.00 & 0.00 & 0.00 \\
accuracy & 0.64 & 0.64 & 0.64 \\
macro avg & 0.63 & 0.64 & 0.61 \\
weighted avg & 0.69 & 0.64 & 0.65 \\
\hline
\end{tabular}
\end{table}

\begin{table}[h]
\centering
\caption{Class-specific performance metrics for GPT 3.5 and GPT 4 in Scenario 3 - Congressional Bills}
\begin{tabular}{|c|c|c|c|}
\hline
Class & Precision & Recall & F1-score \\
\hline
Macroeconomics & 0.66 & 0.45 & 0.53 \\
Civil Rights & 0.72 & 0.66 & 0.69 \\
Health & 0.87 & 0.94 & 0.91 \\
Agriculture & 0.78 & 0.87 & 0.83 \\
Labor & 0.67 & 0.86 & 0.76 \\
Education & 0.83 & 0.92 & 0.87 \\
Environment & 0.73 & 0.82 & 0.77 \\
Energy & 0.92 & 0.84 & 0.88 \\
Immigration & 0.90 & 0.94 & 0.92 \\
Transportation & 0.90 & 0.79 & 0.84 \\
Law and Crime & 0.84 & 0.69 & 0.76 \\
Social Policy & 0.70 & 0.86 & 0.77 \\
Housing & 0.74 & 0.79 & 0.76 \\
Domestic Commerce & 0.54 & 0.61 & 0.57 \\
Defense & 0.85 & 0.39 & 0.54 \\
Technology & 0.81 & 0.79 & 0.80 \\
Foreign Trade & 0.83 & 0.92 & 0.87 \\
International Affairs & 0.75 & 0.27 & 0.40 \\
Government Operations & 0.77 & 0.72 & 0.74 \\
Public Lands & 0.87 & 0.82 & 0.85 \\
Culture & 0.00 & 0.00 & 0.00 \\
Other (Private Bills) & 0.98 & 1.00 & 0.99 \\
\hline
\end{tabular}
\end{table}

\begin{figure}[htp]
\centering
\includegraphics[width=15cm]{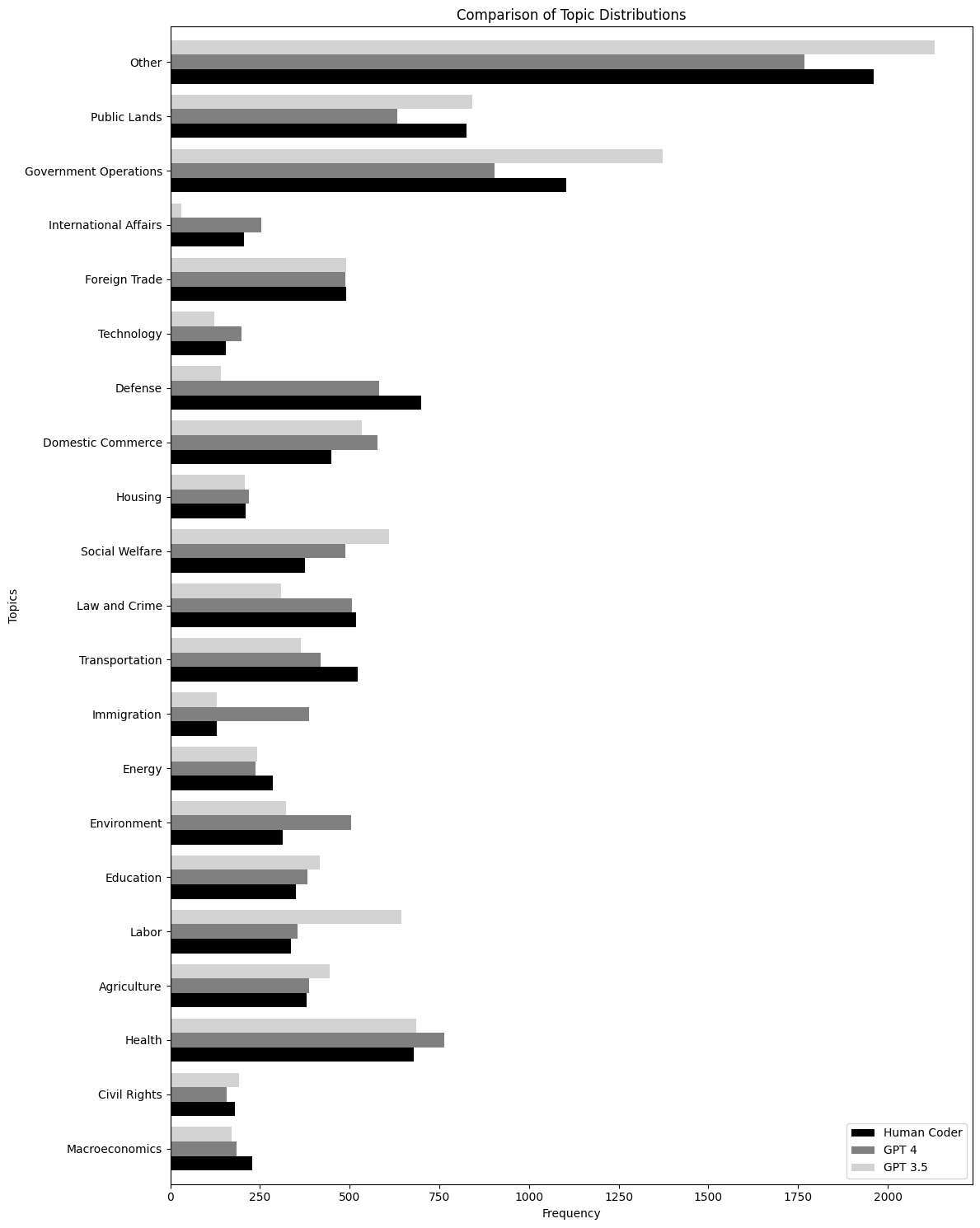}
\caption{Distribution of topics across coding methods - Scenario 2 - Congressional Bills}
\label{fig:image}
\end{figure}

\begin{figure}[htp]
\centering
\includegraphics[width=15cm]{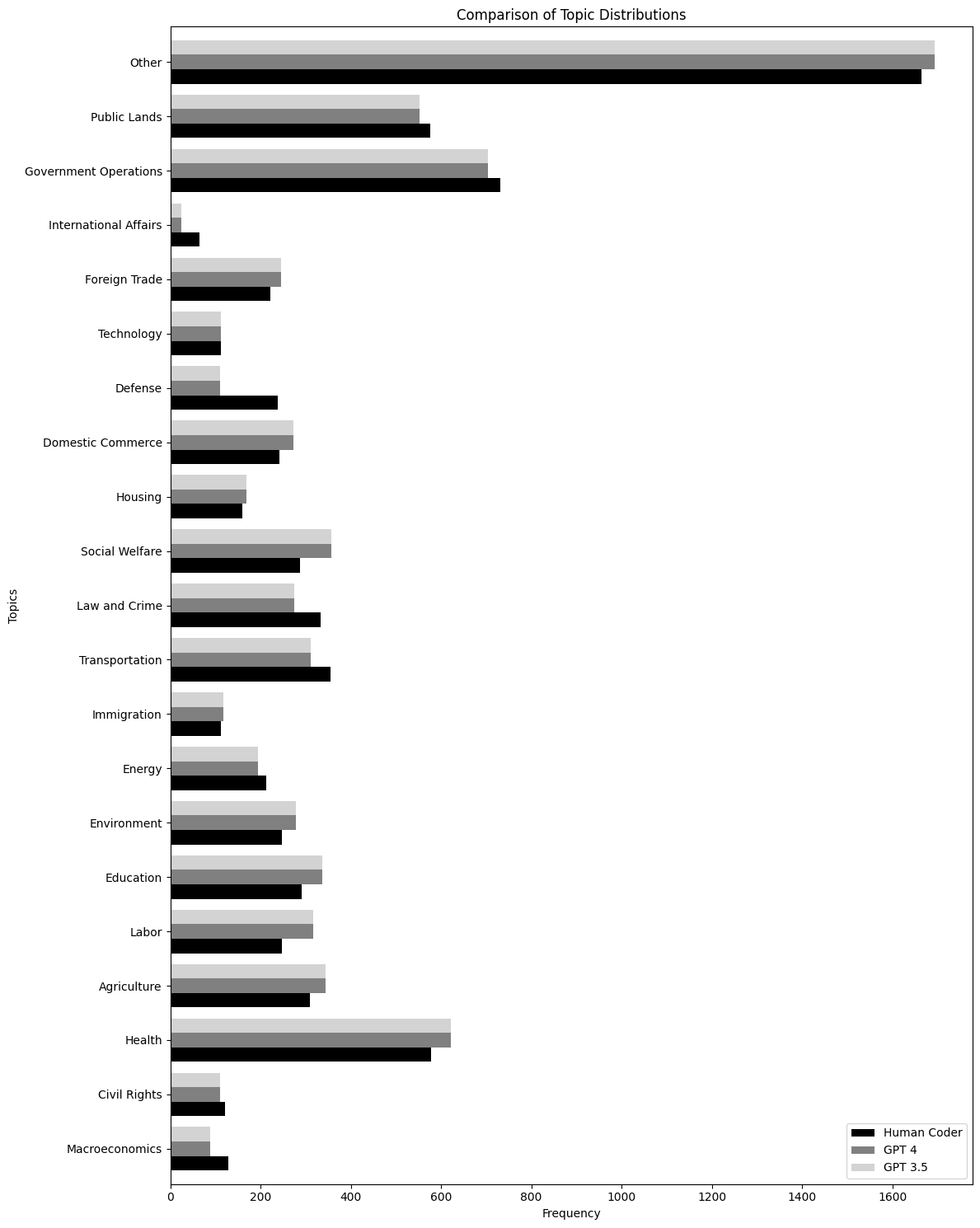}
\caption{Distribution of topics across coding methods - Scenario 3 - Congressional Bills}
\label{fig:image}
\end{figure}

\end{appendices}

\end{document}